\documentclass[10pt,twocolumn,letterpaper]{article}
\newcommand\etc{etc\@ifnextchar.{}{.\@}}

\usepackage{cvpr}

\usepackage{times}
\usepackage{epsfig}
\usepackage{graphicx}
\usepackage{amsmath}
\usepackage{amssymb}
\usepackage{float}
 \usepackage{multirow}
\usepackage{caption} 
\usepackage{hhline}
\usepackage{url}

\makeatletter
\newcommand{\thickhline}{%
    \noalign {\ifnum 0=`}\fi \hrule height 1pt
    \futurelet \reserved@a \@xhline
}
\makeatother

\makeatletter
\renewcommand{\paragraph}{%
  \@startsection{paragraph}{4}%
  {\z@}{1 ex \@plus 0.2ex \@minus .2ex}{-1em}%
  {\normalfont\normalsize\bfseries}%
}
\makeatother

\makeatletter
\g@addto@macro\normalsize{%
  \setlength\abovedisplayskip{4pt}
  \setlength\belowdisplayskip{4pt}
  \setlength\abovedisplayshortskip{4pt}
  \setlength\belowdisplayshortskip{4pt}
}
\makeatother

\usepackage[pagebackref=true,breaklinks=true,letterpaper=true,colorlinks,bookmarks=false]{hyperref}
\cvprfinalcopy 

\def\cvprPaperID{3491} 

\ifcvprfinal\fi
\begin{document}

\title{Skeleton Key: Image Captioning by Skeleton-Attribute Decomposition}

\author{Yufei Wang$^1$  \hspace{0.2cm} Zhe Lin$^2$ \hspace{0.2cm} Xiaohui Shen$^2$
\hspace{0.2cm}   Scott Cohen$^2$
\hspace{0.2cm}   Garrison W. Cottrell$^1$ \\
\and \vspace{-2ex}\\ 
$^1$University of California, San Diego \\
{\tt\small \{yuw176, gary\}@ucsd.edu} \\
\and \vspace{-2ex}\\ 
$^2$Adobe Research \\
{\tt\small \{zlin, xshen, scohen\}@adobe.com}}

\maketitle

\begin{abstract}
Recently, there has been a lot of interest in automatically generating descriptions for an image. Most existing language-model based approaches for this task learn to generate an image description word by word in its original word order. However, for humans, it is more natural to locate the objects and their relationships first, and then elaborate on each object, describing notable attributes. We present a coarse-to-fine method that decomposes the original image description into a skeleton sentence and its attributes, and generates the skeleton sentence and attribute phrases separately. By this decomposition, our method can generate more accurate and novel descriptions than the previous state-of-the-art. Experimental results on the MS-COCO and a larger scale Stock3M datasets show that our algorithm yields consistent improvements across different evaluation metrics, especially on the SPICE metric, which has much higher correlation with human ratings than the conventional metrics. Furthermore, our algorithm can generate descriptions with varied length, benefiting from the separate control of the skeleton and attributes. This enables image description generation that better accommodates user preferences.

\end{abstract}

\section{Introduction}
The task of automatically generating image descriptions, or image captioning, has drawn great attention in the computer vision community. The problem is challenging in that the description generation process requires the understanding of high level image semantics beyond simple object or scene recognition, and the ability to generate a semantically and syntactically correct sentence to describe the important objects, their attributes and relationships.

The image captioning approaches generally fall into three categories. The first category tackles this problem based on retrieval: given a query image, the system searches for visually similar images in a database, finds and transfers the best descriptions from the nearest neighbor captions for the description of the query image \cite{knn, ranking1,treetalk, onemillion}. The second category typically uses template-based methods to generate descriptions that follow predefined syntactic rules\cite{everypicture, babytalk, Li2011, elliott13, yang11, midge}. Most recent work falls into the third category: language model-based methods \cite{fangCVPR15, showandtell, showattend, lrcn2014, mao2014deep, unify2014}. Inspired by the machine translation task \cite{s2s, translation2, translation3}, an image to be described is viewed as a ``sentence" in a source language, and an Encoder-Decoder network is used to translate the input to the target sentence. Unlike machine translation, the source ``sentence" is an image in the captioning task. Therefore, a natural encoder is a Convolutional Neural Network (CNN) instead of a Recurrent Neural Network (RNN).

\begin{figure}[t]
\vspace{-0.1in}
\centering
\makebox[0.5\textwidth][c]{
\includegraphics[width=0.5\textwidth]{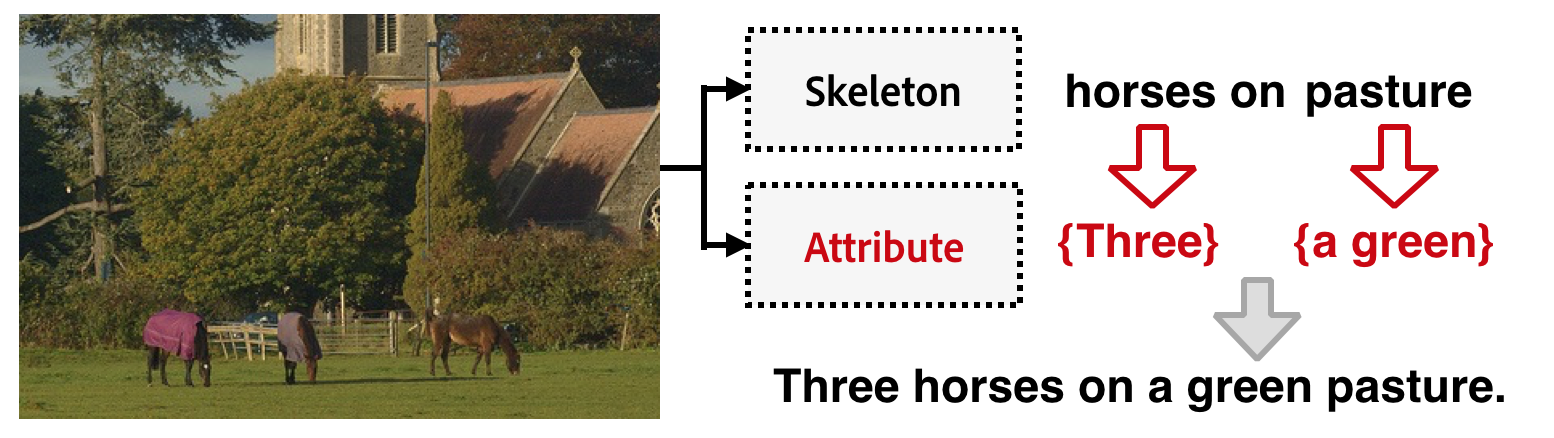}}
\vspace{-0.1in}
\caption{Illustration of the inference stage of our coarse-to-fine captioning algorithm with skeleton-attribute decomposition. First, the skeleton sentence is generated, describing the objects and relationships. Then the objects are revisited and the attributes for each object are generated.}
\label{intro}
\vspace{-0.2in}
\end{figure}

Starting from the basic form of a CNN encoder-RNN decoder, there have been many attempts to improve the system. Inspired by their success in machine translation, Long-short Term Memory (LSTM) networks are used as the decoder in \cite{showandtell, lrcn2014}. Xu \etal \cite{showattend} add an attention mechanism that learns to attend to parts of the image for word prediction. It is also found that feeding high level attributes instead of CNN features yields improvements \cite{semanticAttention, CVPR16What}.

Despite the variation in approaches, most of the existing LSTM-based methods suffer from two problems: 1) they tend to parrot back sentences from the training corpus, and lack variation in the generated captions \cite{quirk}; 2) due to the word-by-word prediction process in sentence generation, attributes  are generated before the object they refer to. Mixtures of attributes, subjects, and relations in a complete sentence create large variations across training samples, which can affect training effectiveness.

In order to overcome these problems, in this paper, we propose a coarse-to-fine algorithm to generate the image description in a two stage manner: First, the skeleton sentence of the image description is generated, containing the main objects involved in the image, and their relationships. Then, the objects are revisited in a second stage using attention, and the attributes for each object are generated if they are worth mentioning. The flow is illustrated in Figure~\ref{intro}. By dealing with the skeleton and attributes separately, the system is able to generate more accurate image captions.

Our work is also inspired by a series of Cognitive Neuroscience studies. During visual processing such as object recognition, two types of mechanisms play important roles: first, a fast subcortical pathway that projects to the frontal lobe does a coarse analysis of the image, categorizing the objects~\cite{bullier01, engel01, gilbert01}, and this provides top-down feedback to a slower, cortical pathway in the ventral temporal lobe ~\cite{tanaka96, bullier95} that proceeds from low level to high level regions to recognize an object. The exact way that the top-down mechanism is involved is not fully understood, but Bar~\cite{bar} proposed a hypothesis that low spatial frequency features trigger the quick ``initial guesses" of the objects, and then the ``initial guesses" are back-projected to low level visual cortex to integrate with the bottom-up process. 

Analogous to this object recognition procedure, our image captioning process also comprises two stages: 1) a quick global prediction of the main objects and their relationship in the image, and 2) an object-wise attribute description. The objects predicted by the first stage are fed back to help the bottom-up attribute generation process. Meanwhile, this idea is also supported by object-based attention theory. Object based attention proposes that the perceptual analysis of the visual input first segments the visual field into separate objects, and then, in a focal attention stage, analyzes a particular object in more detail \cite{neisser1967, object_attention}. 

The main contributions of this paper are as follows: First, we are the first to divide the image caption task such that the skeleton and attributes are predicted separately. Second, our model improves performance consistently against a very strong baseline that outperforms the published state-of-the-art results. The improvement on the recently proposed SPICE \cite{spice2016} evaluation metric is significant.
Third, we also propose a mechanism to generate image descriptions with variable length using a single model. The coarse-to-fine system naturally benefits from this mechanism, with the ability to vary the skeleton/attribute part of the captions separately. This enables us to adapt image description generation according to user preferences, with descriptions containing a varied amount of object/attribute information.

\section{Related Work}

\paragraph{Existing image captioning methods}  Retrieval-based methods  search for visually similar images to the input image, and find the best caption from the retrieved image captions. For example, Devlin \etal  in \cite{knn} propose a K-nearest neighbor approach that finds the caption that best represents the set of candidate captions gathered from neighbor images. This method suffers from an obvious problem that the generated captions are always from an existing caption set, and thus it is unable to generate novel captions.

Template-based methods generate image captions from pre-defined templates, and fill the template with detected objects, scenes and attributes. Farhadi \etal \cite{everypicture} use single $\langle$object, action, scene$\rangle$ triple to represent a caption,  and learns the mapping from images and sentences separately to the triplet meaning space. Kulkarni \etal \cite{babytalk} detect objects and attributes in an image as well as their prepositional relationship, and use a CRF to predict the best structure containing those objects, modifiers and relationships. In \cite{phrase15}, Lebret \etal predict phrases from an image, and combine them with a simple language model to generate the description. 
 These approaches heavily rely on the templates or simple grammars, and so generate rigid captions.

Language model-based methods typically learn the common embedding space of images and captions, and generate novel captions without many rigid syntactical constraints. Kiros and Zemel \cite{icml2014c2_kiros14} propose multimodal log-bilinear models conditioned on image features. Mao \etal \cite{mao2014deep} propose a Multimodal Recurrent Neural Network (MRNN) that uses an RNN to learn the text embedding, and a CNN to learn the image representation. Vinyals \etal \cite{showandtell} use  LSTM as the decoder to generate sentences, and provide the image features as input to the LSTM directly. Xu \etal \cite{showattend} further introduce an attention-based model that can learn where to look while generating corresponding words. You \etal \cite{semanticAttention} use pre-generated semantic concept proposals to guide the caption generation, and learn to selectively attend to those concepts at different time-steps. Similarly, Wu \etal \cite{CVPR16What} also show that high level semantic features can improve the caption generation performance.

Our work is also a language-model-based method.  Unlike approaches to LSTM-based methods that try to feed a better image representation to the language model, we focus on the caption itself, and show how breaking the original word order in a natural way can yield better performance.

 \begin{figure*}[t]
\vspace{-0.2in}
\includegraphics[width=0.8\textwidth]{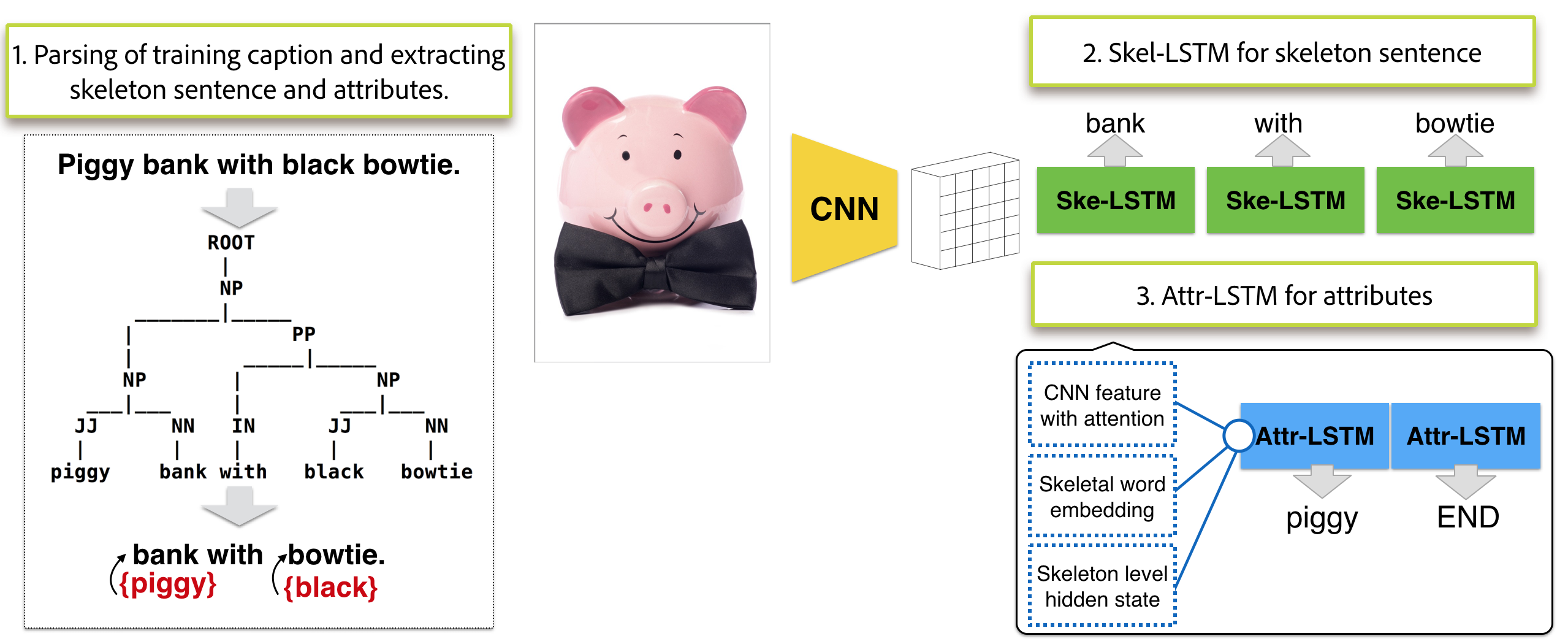}
\centering
\vspace{-0.1in}
\caption{The overall framework of the proposed algorithm. In the training stage, the training image caption is decomposed into the skeleton sentence and corresponding attributes. A Skel-LSTM is trained to generate the skeleton based on the main objects and their relationships in the image, and then an Attr-LSTM generates attributes for each skeletal word.}
\label{architecture}
\vspace{-0.2in}
\end{figure*}

\paragraph{Analyzing the sentences for image captioning}
Parsing of a sentence is the process of analyzing the sentence according to a set of  grammar rules, and generating a rooted parse tree that represents the syntactic structure of the sentence \cite{parser}. There is some language-model-based work that parses the captions for better sentence encoding. For example, Socher \etal \cite{grounded} proposed the Dependency Tree-RNN, which uses dependency trees to embed sentences into a vector space, and then performs caption retrieval with the embedded vector.  Unfortunately, the model is unable to generate novel sentences.

The work that is closest to our own is the hierarchical LSTM model proposed by Tan and Chan \cite{phi}. They view captions as a combination of noun phrases and other words, and try to predict the noun phrases (together with the other words) directly with an LSTM.The noun phrases are encoded into a vector representation with a separate LSTM. In the inference stage, K image-relevant phrases are generated first with the lower level LSTM. Then, the upper level LSTM generates the sentence that contains both the ``noun phrase" token and other words. When a noun phrase is generated, suitable phrases from the phrase pool are selected, and then used as the input to the next time-step. This work is relevant to ours in that it also tries to break the original word order of the caption. However, it directly replaces the phrases with a single word ``phrase token" in the upper level LSTM without distinguishing those tokens, although the phrases can be very different. Also, the phrases in an image are generated ahead of the sentence generation, without knowing the sentence structure or the location to attend to. 

\paragraph{Evaluation metrics}
Evaluation of image caption generation is as challenging as the task itself. Bleu \cite{bleu}, CIDEr \cite{cider}, METEOR \cite{meteor}, and ROUGE \cite{rouge} are common metrics used for evaluating most image captioning benchmarks such as MS-COCO and Flickr30K. However, these metrics are very sensitive to n-gram overlap, which may not necessarily be a good way to measure the quality of an image description. Recently, Anderson \etal \cite{spice} introduced a new evaluation metric called SPICE that overcomes this problem. SPICE uses a graph-based semantic representation to encode the objects, attributes and relationships in the image. They show that SPICE has a much higher correlation with human judgement than the conventional evaluation metrics.

In our work, we evaluate our results using both conventional metrics and the new SPICE metric.
We also show how unimportant words like ``a" impact scores on conventional metrics.

\section{The Proposed Model}
 The overall framework of our model is shown in Figure~\ref{architecture}. In the training stage, the ground-truth captions are decomposed into the skeleton sentences and attributes for the training of two separate networks. In the test stage, the skeleton sentence is generated for a given image, and then attributes conditioned on the skeleton sentence are generated. They are then merged to form the final generated caption.
 
 \subsection{Skeleton-Attribute decomposition for captions}
To extract the skeleton sentence and attributes from a training image caption, we use the Stanford constituency parser \cite{parser, nlpcore}. As shown in Figure~\ref{architecture}, the parser constructs a constituency tree from the original caption, while the nodes hierarchically form phrases of different types. The common phrase types are Noun phrase (NP), Verb phrase (VP), Prepositional phrase (PP), and Adjective phrase (AP). 

To extract the objects in the skeleton sentence, we find the lowest level NP's, and keep the last word within the phrase as the skeletal object word. The words ahead of it within the same NP are attributes describing this skeletal object. The lowest level phrases of other types are kept in the skeleton sentence. 

Sometimes, it is difficult to decide whether all the words except for the last one in a noun phrase are attributes. For example, the phrase ``coffee cup" is a noun-noun compound. Should we keep ``coffee cup" as a single entity, or use ``coffee" as a modifier? In this work, we don't distinguish noun-noun compounds from other attribute-noun word phrases, and treat ``coffee" as the  attribute of ``cup". Our experience is that the  coarse-to-fine network can learn the correspondence, although strictly speaking they are not attribute-object pairs.

\subsection{Coarse-to-fine LSTM}
We use the high level image features extracted from a CNN as the input to the language model. For the decoder part, our coarse-to-fine model consists of two LSTM submodels: one for generating skeleton sentences, and the other for generating attributes. We denote the two submodels as Skel-LSTM and Attr-LSTM respectively.
 

\paragraph{Skel-LSTM} The Skel-LSTM predicts the skeleton sentence given the image features. We adopt the soft attention based LSTM in \cite{showattend} for the Skel-LSTM. Spatial information is maintained in the CNN image features, and an attention map is learned at every time step to focus attention to predict the current word.

We denote the image features at location $(i, j)\in L \times L$ as ${v}_{ij}\in\mathbb{R}^D$. The attention map at time step $t$ is represented as normalized weights $\alpha_{ij,t}$, computed by a multilayer perceptron conditioned on the previous hidden state $h_{t-1}$. 
 \begin{equation}
 \alpha_{ij, t} = \textup{Softmax} (\textup{MLP}({v}_{ij}, h_{t - 1}))
 \end{equation}
Then, the context vector $z_t$ at time $t$ is computed as:
 \begin{equation}
 z_{t} = \sum_{i,j} \alpha_{ij, t} {v}_{ij}
 \end{equation}
 The context vector is then fed to the current time step LSTM unit to predict the upcoming word.
 
Unlike \cite{showattend}, in our model, the attention map $\alpha_{ij, t}$ is not only used to predict the current skeletal word, but also to guide the attribute prediction: the attributes corresponding to a skeletal word describe the same skeletal object, and the attention information we get from Skel-LSTM can be reused in the Attr-LSTM to guide where to look.
 
\paragraph{Attr-LSTM} After the skeleton sentence is generated, the Attr-LSTM predicts the attribute sequence for each skeletal word. Rather than predicting multiple attribute words separately for one object, the Attr-LSTM can predict the attribute sequence as a whole, naturally taking care of the order of attributes. The Attr-LSTM is similar to the model in \cite{showandtell}, with several modifications.

The original input sequence of the LSTM in \cite{showandtell} is:
\begin{equation}
x_{-1} = \textup{CNN}(I)
\end{equation}
\begin{equation}
x_{t} = W_e y_t, t = 0, 1, ..., N - 1
\label{eq1}
\end{equation}
where $I$ is the image, $\textup{CNN}(I)$ is the CNN image features as a vector without spatial information, $W_e$ is the learned word embedding, and $y_t$ is the ground-truth word encoded as a one-hot vector. $y_0$ is a special start-word token.

In our coarse-to-fine framework, attribute generation is conditioned on the skeletal word it is describing. Therefore, apart from the image features, the Attr-LSTM should be informed by the current skeletal word. On the other hand, the context of the skeleton sentence is also important to give the Attr-LSTM a global understanding of the caption, rather than just focusing on the single current skeletal word. We experimented with feeding the skeletal hidden activations from different time steps into the Attr-LSTM, including the previous time step, the current time step, and the final time step, and found that the current time step hidden activations yield the best result. Moreover, as mentioned in Skel-LSTM, rather than using global image features as the input, we use attention-based image features to encourage the attribute predictor to focus on the current skeletal word.

We formulate the input of Attr-LSTM at the first time step as a multilayer network that fuses different sources of information into the embedding space:
\begin{equation}
x_{-1} = \textup{MLP}(W_I z_{T} + W_t s^{skel}_T + W_h h^{skel}_{T})
\end{equation}
where $T$ is the time step of the current skeletal word, 
$z_{T}  \in \mathbb{R}^{D}$ is the attention weighted average of the image features, 
 $s^{skel}_T \in \mathbb{R}^{m_s}$ is the embedding of the skeletal word at time $T$,
 $h^{skel}_T \in \mathbb{R}^{n_s}$ is the hidden state in the Skel-LSTM of dimension $n_s$. $m_s$ and $n_s$ are dimensionality of the Skel-LSTM word embedding, and the LSTM units, respectively.
$W_l, W_t, W_h$ are learned parameters. The remaining input to Attr-LSTM is the same as Equation~\ref{eq1}. The Attr-LSTM framework is illustrated in Figure~\ref{architecture}. 

In the training stage, the ground truth skeleton sentence is fed into the Skel-LSTM, and $s^{skel}_T$ is the ground truth skeleton word embedding. In test stage, $s^{skel}_T$ is the embedding of predicted skeleton word.

\paragraph{Attention refinement for attribute prediction} Optionally, we can refine the attention map acquired in the Skel-LSTM for better localization of the skeletal word, thus improving the attribute prediction. The attention map $\alpha$ is a pre-word $\alpha$ that is generated before the word is predicted. It can cover multiple objects, or can even be in a different location from the predicted word. Therefore, a refinement of the attention map after the prediction of the current word can provide more accurate guidance for the attribute prediction.

\begin{figure}[t]
\makebox[0.5\textwidth][c]{
\includegraphics[width=0.5\textwidth]{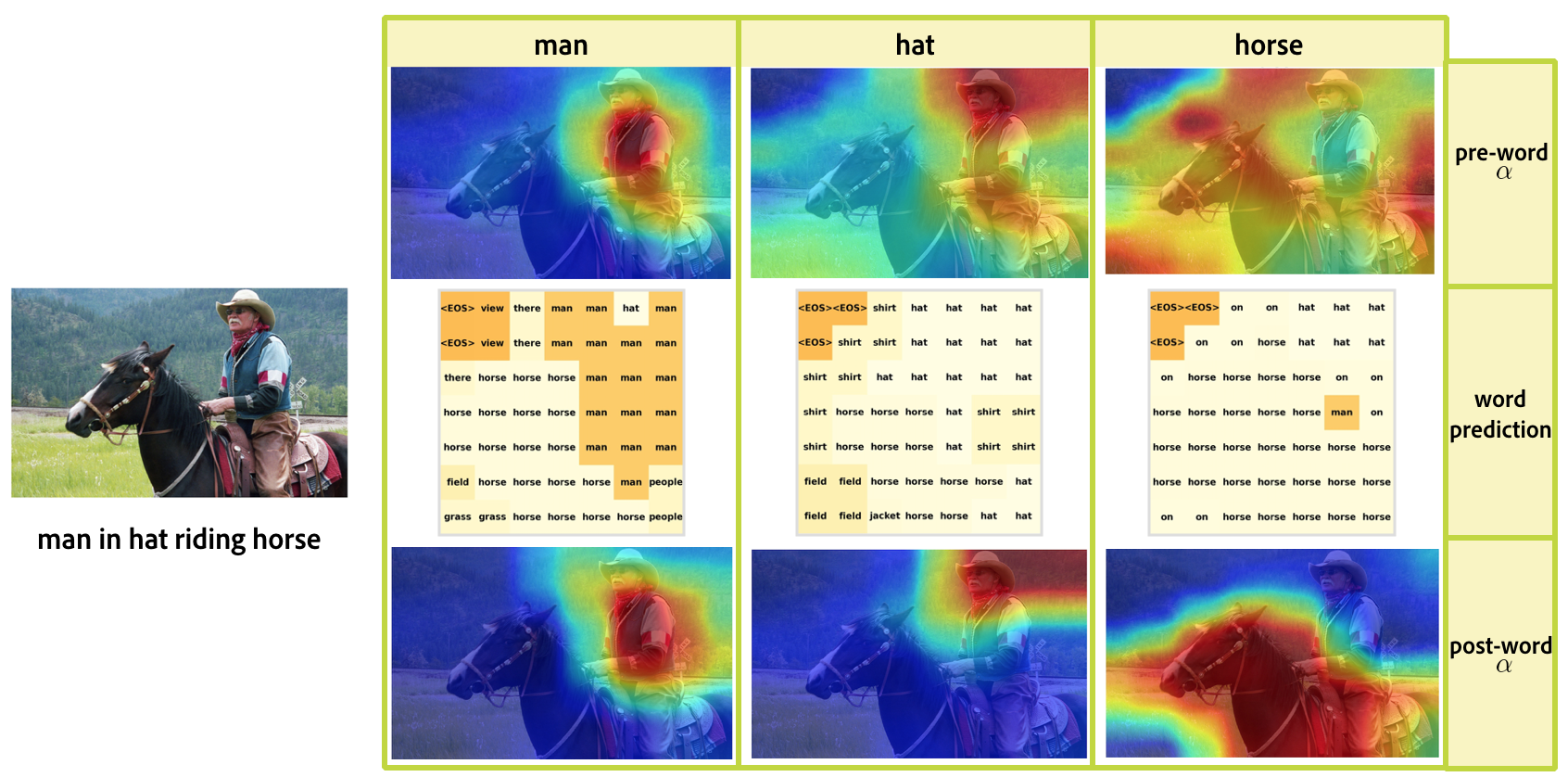}}
\centering
\caption{Illustration of attention refinement process. Due to limited space, only three object words are shown from the predicted caption ``man in hat riding horse". For each word, the attention map, predicted words for each location, and refined attention map are shown. We provide more examples in the supplementary material. }
\vspace{-0.1in}
\label{attention}
\vspace{-0.1in}
\end{figure}

The LSTM unit at time step $T$ outputs the word probability prediction $P_{attend} = (p_1, p_2, ..., p_Q)$, where $Q$ is the vocabulary size in Skel-LSTM. In addition to the single weighted sum feature vector $z_{T}$, we can also use the feature vector $v_{ij}$ in each location as input to the Skel-LSTM. Thus, for each of the $L^2$ locations, we can get the probability of word prediction $P_{ij}$. We can use the spatial word probability to refine the attention map $\alpha$:
\begin{equation}
\alpha_{post(ij)} = \frac{1}{Z}P^T_{attend} \cdot P_{ij}
\end{equation}
where $Z$ is the normalization factor so that $\alpha_{post(ij)}$ sums to one. The refined post-word $\alpha$ is proportional to the similarity between $P_{attend}$ and $P_{ij}$. In Figure~\ref{attention}, we illustrate the attention refinement  process.

\paragraph{Fusion of Skeleton-Attributes} After attributes are predicted for all the skeletal words, attributes are merged into the skeleton sentence just before the corresponding skeletal word, and the final caption is formed.
 \subsection{Variable-length caption generation}
 \label{suppression}
 Due to the imperfections in the current parser approach that we use, there are some cases where the parsing result is noisy. Most of the time, the noise is from incorrect noun phrase recognition, and short skeleton sentences with one or several missing objects.
 This leads to a shorter skeleton prediction in the Skel-LSTM on average, thus eventually causes shorter predictions for the full sentence.
 
To overcome this problem, we designed a simple yet effective trick to vary the length of the generated sentence. Without modifying the trained network, 
In the inference stage of either Skel-LSTM or Attr-LSTM, 
we modify the sentence probability with a length factor:
 \begin{equation}
\textup{log}(\hat{P}) = \textup{log}(P) + \gamma \cdot l
\end{equation}
Where $P$ is the probability of a generated sentence, and $\hat{P}$ is the modified sentence probability. $l$ is the length of the generated sentence. $\gamma$ is the length factor to encourage or discourage longer sentences. Note that the modification is performed during generation of the each word rather than performed after the whole sentence is generated. It is equivalent to adding $\gamma$ to each word log probability except for the end-of-sentence token $\langle$EOS$\rangle$ when sampling the next word from the word probability distribution. This trick of sentence probability modification works well together with beam search.

 
 Our coarse-to-fine algorithm especially benefits from this mechanism, since it can be applied to either Skel-LSTM or Attr-LSTM, resulting in varied information in either objects, or the description of those objects. This allows us to generate captions according to user preference on the complexity of captions and amount of information in captions.

 \section{Experiments}
In this section,  we describe our experiments on two datasets to test our proposed approach.

 \subsection{Datasets}
 We perform experiments on two datasets: the popular benchmark MS-COCO, and Stock3M, a new dataset with much larger scale and more natural captions. 
 
MS-COCO has 123,287 images. Each image is annotated with 5 human generated captions, with an average length of 10.36 words. We use the standard training/test/validation split that is commonly used by other work \cite{semanticAttention, CVPR16What}, and use 5000 images for testing, and 5000 images for validation.

 
MS-COCO is a commonly used benchmark for image captioning tasks. However, there are some issues with the dataset: the images are limited and biased to certain content categories, and the image set is relatively small. Moreover, the captions generated by AMT workers are not particularly natural.  Therefore, we collected a new dataset: Stock3M. Storck3M contains 3,217,654 user uploaded images with a large variety of content. Each image is associated with one caption that is provided by the photo uploader on a stock website. 
 The caption given by the photo uploader is more natural than those found in MS-COCO, and the dataset is 26 times larger in terms of number of images. The captions are much shorter than MS-COCO, with an average length of 5.25 words, but they are more challenging, due to a larger vocabulary and image content variety. We use 2000 images for validation and 8000 images for testing.
 
  \subsection{Experimental details}
  \paragraph{Preprocessing of captions} We follow the preprocessing procedure in \cite{karpathy15} for the captions, removing the punctuation and converting all characters to lower case. For MS-COCO, we discard words that occur fewer than 5 times in skeleton sentences, and fewer than 3 times in attributes. This results in 7896 skeleton, and 5199 attribute words. In total, there are 9535 unique words. For the baseline method that processes the full sentences, a similar preprocessing procedure is applied to the full sentences. Words that occur less than 5 times are discarded, resulting in 9567 unique words.

For Stock3M, due to the larger vocabulary size, we set the word occurrence thresholds to 30 for skeleton and 5 for attributes respectively. This results in 11047 skeleton and 12385 attribute words, with a total of 14290 unique words. In the baseline method that processes full sentences, the occurrence threshold is 30, resulting in 13788 unique words.

\paragraph{Image features and training details for MS-COCO } It has been argued that high level features such as attributes are better as input to caption-generating LSTMs \cite{semanticAttention, CVPR16What}. Our empirical finding is that by simply adopting a better network architecture that provides better image features, and fine-tuning the CNN within the caption dataset, the features extracted are already excellent inputs to the LSTM.
We use ResNet-200 \cite{he15deepresidual} as the encoder model. Images are resized to $256 \times 256$ and randomly cropped to $224 \times 224$. The layer before the average pooling layer and classification layer is used for the image features. and it outputs features with size $2048 \times 7 \times 7$, maintaining the spatial information.

Our system is implemented in Torch \cite{torch}. We fine-tune the CNN features as follows: first, the CNN features are fixed, and an LSTM is trained for full sentence generation. After the LSTM achieves reasonable results, we start fine-tuning the CNN with learning rate 1e-5. The fine-tuned CNN is then used for both Skel-LSTM and Attr-LSTM.
The parameters for the Decoder network are as follows: word embedding is trained from scratch, with a dimension of 512. For Skel-LSTM, we set the learning rate 0.0001, and the hidden layer dimension 1800. For Attr-LSTM, the learning rate is 0.0004, and the hidden layer is 1024-dimensional.  Adagrad is used for training. The learning rate is cut in half once after the validation loss stops dropping.

\paragraph{Image features and training details for Stock3M } We use GoogleNet \cite{googlenet} fine-tuned on Stock3M as the CNN encoder, and add an embedding module after the 1024-dimensional output of GoogleNet $pool5/7\times 7s1$ layer. 

Stock3M is different from MS-COCO in that the images mostly contain single objects, and the captions are more concise than MS-COCO. The average length of Stock3M captions is about half that of MS-COCO. Hence, we did not observe improvement with the attention mechanism, because there are fewer things to focus on. For simplicity, we use the LSTM in \cite{showandtell} for Skel-LSTM. Consequently, for Attr-LSTM, there is no attention input in the -1 time step. We will show that even without attention, the coarse-to-fine algorithm improves substantially over baseline.

\paragraph{Parameters in the testing stage} For both Skel-LSTM and Attr-LSTM, we use a beam search strategy, and adopt length factor $\gamma$ as explained in Section~\ref{suppression}. The beam size and value of $\gamma$ are chosen using the validation set, and are provided in supplementary material.

\begin{table*}[t]
\centering
\caption{Performance of our proposed method and the baseline method on SPICE measurement, for the two datasets. We also include the results on different semantic concept subcategories.}
\scalebox{0.9}{
\begin{tabular}{c|c|ccccccccc}
\thickhline
                                              & \multicolumn{1}{c|}{\textbf{Model}}    & \multicolumn{1}{c}{\textbf{SPICE}} & \multicolumn{1}{c}{\textbf{Precision}} & \multicolumn{1}{c}{\textbf{Recall}} & \multicolumn{1}{c}{\textbf{Object}} & \multicolumn{1}{c}{\textbf{Relation}} & \multicolumn{1}{c}{\textbf{Attribute}} & \multicolumn{1}{c}{\textbf{Size}}  & \multicolumn{1}{c}{\textbf{Color}} & \multicolumn{1}{c}{\textbf{Cardinality}} \\ \hline
\multirow{2}{*}{\textbf{MS-COCO}}                    & {Baseline  }    &                      0.188            &           0.508                &           0.117                &                   0.350         &     0.048                         &                   0.098              &           0.045                    &          0.132                 &                  0.039                   \\
                                               & {Ours  }    &                      \textbf{0.196}            &           \textbf{0.529}                &           \textbf{0.123}                  &                   \textbf{0.363}         &     \textbf{0.050}                         &                   \textbf{0.110 }             &           \textbf{0.073}                    &          \textbf{0.170 }                &                  \textbf{0.064 }                  \\\hline
\multirow{2}{*}{\textbf{Stock3M}  }                    & {Baseline}                      & 0.157                     & 0.173                         & 0.166                      & 0.250                       & 0.049                           & 0.077                         & 0.129                     & 0.135                     &  -                      \\
                                              & {Ours}                          & \textbf{0.172}                     & \textbf{0.190}                         & \textbf{0.185}                      & \textbf{0.276}                        & \textbf{0.061}                           & \textbf{0.081}                         & \textbf{0.144}                     & \textbf{0.151}                     &  -                     
\\ \thickhline
\end{tabular}
}
\label{table1}
\end{table*}

 \subsection{Results}
 \paragraph{Evaluation metrics} Apart from the conventional evaluation metrics that are commonly used: Bleu \cite{bleu}, CIDEr \cite{cider}, METEOR\cite{meteor}, and ROUGE \cite{rouge}, we use the recently proposed SPICE metric \cite{spice}, which is not sensitive to n-grams and builds a scene graph from captions to encode the objects, attributes and relationships in the image. We emphasize our performance on this metric, because it has much higher correlation with human ratings than the other conventional metrics, and it shows the performance specific to different types of information, such as different types of attributes, objects, and relationships between objects. 
 \paragraph{Baseline} In order to demonstrate the effectiveness of our method, we also present a baseline result. The baseline method is trained and tested on full caption sentences, without skeleton-attribute decomposition. For each dataset, we use the same network architecture as in the Skel-LSTM architecture, and use the same hyper-parameters and the same CNN encoder as in our proposed coarse-to-fine method.
\paragraph{Quantitative results}
We report both SPICE in Table~\ref{table1} and conventional evaluation metrics in Table~\ref{table2}.

\begin{table*}[t]
\centering
\caption{Performance of our proposed methods and other state-of-the-art methods on MS-COCO and Stock3M. Only scores that were reported in the papers are shown here.}
\label{table2}
\scalebox{0.9}{
\begin{tabular}{c|c|ccccccc}
\thickhline
\textbf{Datasets}                 & \textbf{Models}   &  \textbf{B-1}   &  \textbf{B-2}   &  \textbf{B-3}   &  \textbf{B-4}   &  \textbf{METEOR} &  \textbf{ROUGE-L} &  \textbf{CIDEr} \\ \hline
\multirow{9}{*}{\textbf{MS-COCO}} & NIC  \cite{showandtell}    & -     & -     & -     & 0.277 & 0.237  & -       & 0.855 \\
                         & LRCN  \cite{lrcn2014}   & 0.669 & 0.489 & 0.349 & 0.249 & -      & -       & -     \\
                         & Toronto \cite{showattend}  & 0.718 & 0.504 & 0.357 & 0.250 & 0.230  & -       & -     \\
                         & ATT  \cite{semanticAttention}    & 0.709 & 0.537 & 0.402 & 0.304 & 0.243  & -       & -     \\
                         & ACVT  \cite{CVPR16What}   & 0.74  & 0.56  & 0.42  & 0.31  & 0.26   & -       & 0.94  \\ \cline{2-9} 
                         & Baseline &     0.742   &  0.577     &    0.442   &    0.340   &    0.268    &    0.552     &   1.069        \\
                         & Ours     &     0.742    &   0.577    &    0.440   &   0.336    &   0.268     &   0.552      &   1.073     \\ \hhline{~========}
                         & Baseline (w/o \textit{a})&    0.664    &   0.481    &    0.351   &   {0.258}    &     0.245   &    0.485     &  0.949        \\
                         & Ours  (w/o \textit{a})   &     \textbf{0.673}    &    \textbf{0.489}   &  \textbf{0.355}     &  \textbf{0.259}     &    \textbf{0.247}    &     \textbf{0.489 }   &     \textbf{0.966}        \\ \thickhline \hline
\multirow{4}{*}{\textbf{Stock3M}} & Baseline &    0.236   &     0.133  &    0.079   &     0.050  &     0.108   &   0.233      &   0.720    \\
                         & Ours     &   \textbf{0.245}    &    \textbf{0.138}   &    \textbf{0.083}   &     \textbf{0.052}  &    \textbf{0.110}     &    \textbf{0.239}     &   \textbf{0.724}    \\ \hhline{~========}
                         & Baseline (w/o \textit{a})&    0.233   &  0.134     &  0.082     &   0.053    &    0.108    &     0.235    &  0.737     \\
                         & Ours (w/o \textit{a})    &   \textbf{0.246}    &    \textbf{0.140}   &  \textbf{0.086}     &    \textbf{0.055}   &     \textbf{0.111}   &    \textbf{0.241}     & \textbf{0.738}      \\ \thickhline
\end{tabular}
}
\vspace{-0.15in}
\end{table*}

First, it is worth noting that our baseline method is a very strong baseline. In Table~\ref{table2}, we compare our method with  published state-of-the-art methods. Our baseline method already outperforms the state-of-the-art by a considerable margin, indicating the importance of a powerful image feature extractor. 
By just fine-tuning the CNN with the simple baseline algorithm, we outperform the approaches with augmentation of high level attributes \cite{semanticAttention, CVPR16What}. The baseline already ranks 3rd - 4th place on the MS-COCO CodaLab leaderboard\footnote{\url{https://competitions.codalab.org/competitions/3221}}. Note that we use no augmentation tricks such as ensembling, or scheduled sampling \cite{lessons}, which can improve the performance further. We provide our submission to the leaderboard in the supplementary material.


SPICE is an F-score of the matching tuples in predicted and reference scene graphs. It can be divided into meaningful subcategories. In Table~\ref{table1} we report the SPICE score as well as the subclass scores of objects, relations and attributes. In particular, size, color, and count attributes are reported. Table~\ref{table1} shows consistent improvement over  baseline for the two datasets, and this extends to the subcategories. The cardinality F-score for Stock3M is not reported here because there are too few images with this type of attribute to have a meaningful evaluation: there are only 78 cardinality attributes out of 8000 test images.

In Table~\ref{table2}, we also show the comparison between the proposed method and baseline method on conventional evaluation metrics. As shown, there is no significant improvement over baseline on most of the conventional metrics on MS-COCO. This is due to an intrinsic problem with the conventional metrics: they overly rely on n-gram matching. The proposed coarse-to-fine algorithm breaks the original word order of the training captions, and thus weakens the objective of predicting exact n-grams as in the training captions. There is even a small drop on BLEU-3 and BLEU-4 on MS-COCO against the baseline. To investigate if the two methods indeed have similar performance as reflected in those conventional metrics, we conducted further analysis of the results.

We preprocess the ground-truth and predicted captions to remove all the \textit{a}'s in the captions. This is motivated by the observation that 15\% of words in the MS-COCO captions are \textit{a}. This function word affects the n-gram match greatly, though it conveys very little information in the MS-COCO like captions. Therefore, by removing the \textit{a}'s in the captions,  we obtain a measure that is not influenced by the n-grams using \textit{a}, and hence is more focused on content words. The performance evaluation on the same datasets with \textit{a} removed is shown in Table~\ref{table2} as  ``Baseline/Ours (w/o \textit{a})". It can be seen that consistent improvement is achieved with our coarse-to-fine method.

In Table~\ref{table2}, we also present the performance of our coarse-to-fine method as well as the baseline method on Stock3M  evaluated on conventional metrics. In Stock3M,  the frequency of the word \textit{a} is only 2.5\%, therefore it has no big impact on the relative performance of the two methods. We can see consistent improvement on all the metrics.


\subsection{Analysis of generated descriptions}
\label{analysis}

\begin{figure*}[t]
\vspace{-0.15in}
\includegraphics[width=0.9\textwidth]{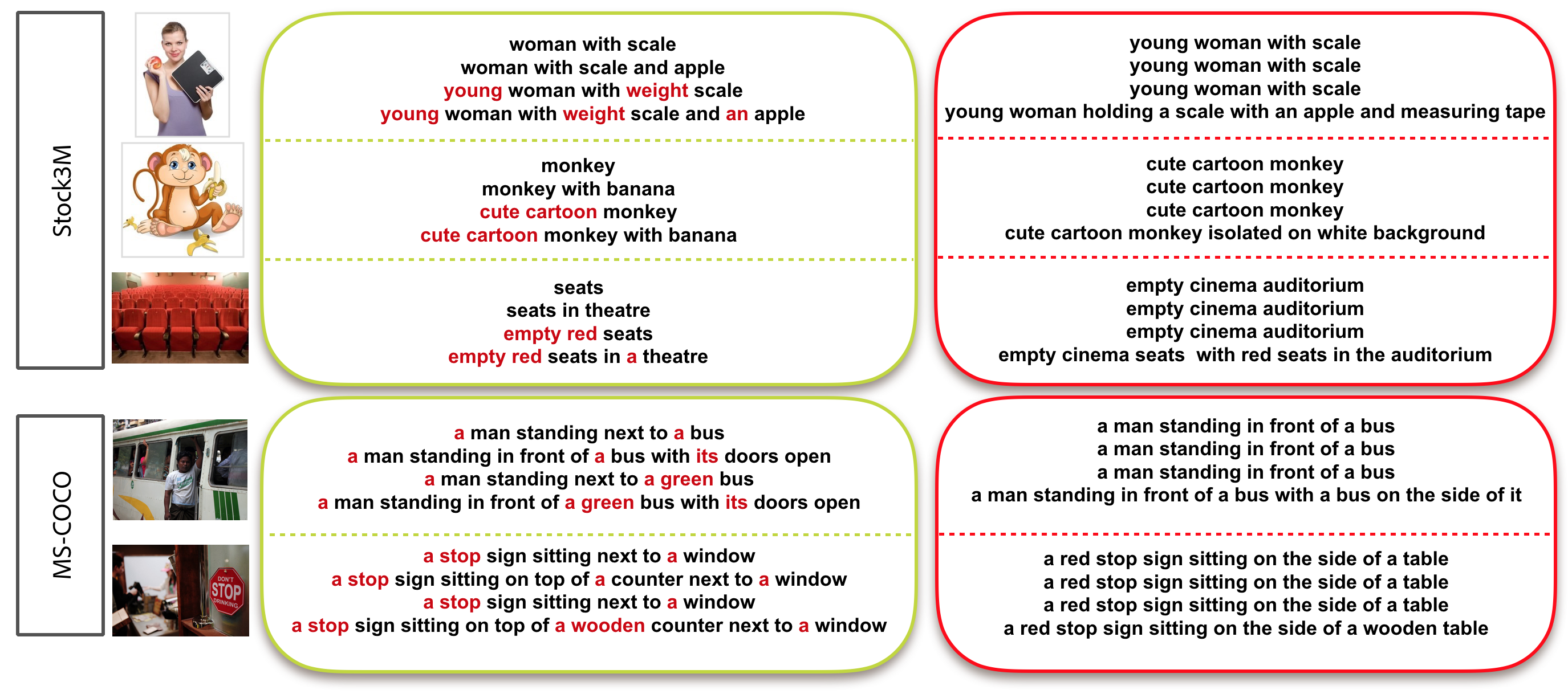}
\centering
\vspace{-0.15in}
\caption{Examples of predicted titles for image examples from Stock3M and MS-COCO. Four titles are generated from our coarse-to-fine model (middle, in green box) and baseline model (right, in red box) respectively. For the coarse-to-fine model, four pairs of length factor value $\gamma$ for skeletal title and attributes are (-1, -1), (1.5, -1),(-1, 1.5),  (1.5, 1.5) respectively. For the baseline method, the $\gamma$'s are -1, -0.5, 0.5, 1.5 respectively.}
\label{varied}
\vspace{-0.17in}
\end{figure*}

\paragraph{Generating variable-length captions.}
In the coarse-to-fine algorithm, a length factor is applied to the Skel-LSTM and Attr-LSTM separately to encourage longer skeleton/attribute generation in order to generate captions that have similar length to the training captions. However, we can further manually tune the length factor value to control the length of skeleton/attribute of the generated captions. In Figure~\ref{varied}, we show some test examples from Stock3M  and MS-COCO . For each of the images, four captions are generated with four pairs of (skeleton, attribute) length factor values: (-1, -1), (1.5, -1), (-1, 1.5),  (1.5, 1.5). The four value pairs represent all combinations of encouraging less/more information in skeleton/attributes. Attributes are marked in red in the generated caption. We can see how the length factor works together with beam search to get syntactically and semantically correct captions. The amount of object/attribute information naturally varies with the length of the skeleton/attributes. 

We can certainly apply the same trick on the baseline method using different length factor values. For comparison, in Figure~\ref{varied} (red box), we show the four captions generated from baseline method also using four different length factor values: $\gamma \in \left \{ -1, -0.5, 0.5, 1.5 \right \}$. As illustrated, although the captions generated by the baseline model can also have different lengths, they are much less flexible and useful than the ones generated by our coarse-to-fine model. This is because the coarse-to-fine model can decompose the caption into skeletons and attributes, and have separate requirements for objects and attributes according to user preference: the user may prefer descriptions that only describe main objects but in more detail; or he/she may prefer descriptions that contain all the objects in the images, but cares less about the object attributes.

\paragraph{Post-word $\alpha$ helps with attribute prediction}
The results we show in Table~\ref{table1} and \ref{table2} for the proposed coarse-to-fine model adopts attention refinement for attribute prediction in the Attr-LSTM on MS-COCO. Here, we further validate the effectiveness of the post-word $\alpha$ refinement approach in Table~\ref{refinement}, by comparing the result without attention refinement ({Pre-word $\alpha$}) with the result with attention refinement ({Post-word $\alpha$}). The post-word $\alpha$ only refines the attended area for attribute prediction, therefore we only show the improvement of SPICE score on attribute subcategories. The performance on other categories is unchanged. We see consistent improvement across different types of attributes, especially on color and size. This shows that a good attention map can improve attribute prediction.

\begin{table}[t]
\caption{Comparison of our proposed method with and without post-word $\alpha$ attention on MS-COCO.}
\scalebox{0.9}{
\centering
\label{refinement}
\begin{tabular}{c|cccc}
\hline
\textbf{Model}          & \textbf{Attribute} & \textbf{Color} &  \textbf{Size} & \textbf{Cardinality}\\ \hline
Pre-word $\alpha$       & 0.107       & 0.167     & 0.069  & 0.063 \\
Post-word $\alpha$ & \textbf{0.110}         & \textbf{0.170}   &  \textbf{0.073} &   \textbf{0.064} \\\hline
\end{tabular}
}
\vspace{-0.15in}
\end{table}

\paragraph{The ability to generate unique and novel captions}
It has been pointed out that the current LSTM based method has a problem generating sentences that have not been seen in the training set, and generates the same sentences for different test images \cite{quirk}. This means that the LSTM dynamics are caught in a rut of repeating the sequences it was trained on for visually similar test images, and is less capable of generating unique sentences for a new image with an object/attribute composition that is not seen in the training set. With the skeleton-attribute decomposition, we claim that our algorithm can generate more unique captions, and can give more accurate attributes even when the attribute-object pattern is new to the system. As shown in Table~\ref{unique_table}, our coarse-to-fine model increases the percentage of generated unique captions by 3\%, and increases the percentage of novel captions by 8\%.

\begin{table}[t]
\caption{Percentage of generated unique sentences and captions seen in training captions for the baseline method and our coarse-to-fine method. The statistics are gathered from the test set of MS-COCO  containing 5000 images.}
\scalebox{0.9}{
\centering
\label{unique_table}
\begin{tabular}{c|cc}
\hline
\textbf{Model}          & \textbf{Unique captions} & \textbf{Seen in training} \\ \hline
Baseline       & 63.96\%         & 56.06\%          \\
Coarse-to-fine & 66.96\%         & 47.76\%         \\\hline
\end{tabular}
}
\vspace{-0.15in}
\end{table}

\paragraph{Qualitative result of generated captions} In the supplementary material, we show more qualitative examples of generated captions from our coarse-to-fine model and baseline model.

\section{Conclusion}
In this paper, we propose a coarse-to-fine model for image
caption generation. The proposed model decomposes
the original image caption into a skeleton sentence and corresponding
attributes, and formulates the captioning process in a natural
way in which the skeleton sentence is generated first, and then
the objects in the skeleton are revisited for attribute generation.
We show with experiments on two challenging
datasets that the coarse-to-fine model can generate better
and more unique captions over a strong baseline method.
Our proposed model can also generate descriptive captions
with variable lengths separately for skeleton sentence and
attributes, and this allows for caption generation according
to user preference.

In future work, we plan to investigate
more complicated skeleton/attribute decomposition
approaches, and allow for attributes that appear after the
skeletal object. It is also of interest  to design a model
that automatically decides on the length of generated
caption based on the visual complexity of the image.

{\small
\bibliographystyle{ieee}
\bibliography{egbib}
}

\end{document}